# Semi-Supervised and Self-Supervised Collaborative Learning for Prostate 3D MR Image Segmentation

*Yousuf Babiker M. Osman[1,2#], Cheng Li[1#], Weijian Huang[1,2,3], Nazik Elsayed[1,2,5], Zhenzhen Xue[1], Hairong Zheng[1], Shanshan Wang\*[1,3,4]*

1 Paul C. Lauterbur Research Center for Biomedical Imaging, Shenzhen Institute of Advanced Technology, Chinese Academy of Sciences, Shenzhen 518055, China.

2 University of Chinese Academy of Sciences, Beijing 100049, China.

3 Peng Cheng Laboratory, Shenzhen 518066, China.

4 Guangdong Provincial Key Laboratory of Artificial Intelligence in Medical Image Analysis and Application, Guangzhou 510080, China.

5 Faculty of Mathematical and Computer Sciences, University of Gezira, Wad Madani, Sudan

**ABSTRACT**

Volumetric magnetic resonance (MR) image segmentation plays an important role in many clinical applications. Deep learning (DL) has recently achieved state-of-the-art or even human-level performance on various image segmentation tasks. Nevertheless, manually annotating volumetric MR images for DL model training is labor-exhaustive and time-consuming. In this work, we aim to train a semi-supervised and self-supervised collaborative learning framework for prostate 3D MR image segmentation while using extremely sparse annotations, for which the ground truth annotations are provided for just the central slice of each volumetric MR image. Specifically, semi-supervised learning and self-supervised learning methods are used to generate two independent sets of pseudo labels. These pseudo labels are then fused by Boolean operation to extract a more confident pseudo label set. The images with either manual or network self-generated labels are then employed to train a segmentation model for target volume extraction. Experimental results on a publicly available prostate MR image dataset demonstrate that, while requiring significantly less annotation effort, our framework generates very encouraging segmentation results. The proposed framework is very useful in clinical applications when training data with dense annotations are difficult to obtain.

## 1. INTRODUCTION

Accurate prostate magnetic resonance (MR) image segmentation is crucial for the diagnosis and treatment of prostate diseases, specifically prostate cancer, one of the most widespread cancers affecting the healthcare of men [1]. However, manually segmenting volumetric MR images is labor-exhaustive and time-consuming. Errors might occur because of fatigue. As a result, high-performance automated prostate segmentation algorithms are increasingly demanded in routine clinical practice to achieve fast and accurate computer-aided diagnoses.

Deep learning-based medical image segmentation has been very popular lately, which can be roughly categorized into fully-supervised, semi-supervised and unsupervised learning. Recent years have seen an exponential increase in the amount of research being done on fully-supervised learning for medical image segmentation [2]. A large amount of densely labeled data is essential for the success of most of these approaches [3]. One recently widely utilized framework in this category is nnU-Net [4]. Nevertheless, collecting enough annotated 3D medical data is a tedious and expensive process. Consequently, relieving this burden is extremely beneficial [5].

Semi-supervised learning methods, which exploit the information of both labeled and unlabeled data, have been extensively investigated to reduce annotating effort. One simple and effective semi-supervised learning approach is to generate pseudo labels for unlabeled data [6, 7]. However, the current prevalent semi-supervised learning tasks address the challenge of having a small amount of voxel-wise densely annotated data with a large amount of unlabeled data. Learning with only single slice annotation for volumetric medical image segmentation is far under-investigated. Existing approaches mainly tackle this problem by label propagating, whereas there is this issue of error accumulation that needs to be addressed carefully [8]. It is still an open question on how to generate reliable pseudo labels.

Unsupervised learning seems to be a very fascinating research direction. A typical area is self-supervised learning. Contrastive learning, which is a popular form of self-supervised learning, has been successfully employed in medical image segmentation [9]. Feature clustering is another feasible unsupervised learning approach [10]. In general, most of existing unsupervised learning methods for medical image segmentation either rely on a small labeled dataset to fine-tune the model or have limited performance when compared to fully-supervised or semi-supervised learning approaches.

In this work, we tackle the challenge of volumetric prostate 3D MR image segmentation utilizing extremely sparse annotations, for which the ground truth manual labels are only available for the central slice of each 3D training image. Specifically, we develop a semi-supervised and self-supervised collaborative learning framework to generate more reliable pseudo labels for information exploitation from unlabeled slices. To fuse the pseudo labels generated by the two learning approaches methods, an effective Boolean operation is adopted. Our final segmentation model is trained using data with either manual labels or network self-generated pseudo labels. The proposed framework generates very encouraging prostate segmentation results when compared to existing methods.

## 2. METHODOLOGY

The three components of our proposed approach — pseudo label generation, pseudo label fusion, and target segmentation network training — are shown in Figure 1.

#These authors contributed equally to this work.
*Corresponding author: sophiasswang@hotmail.com

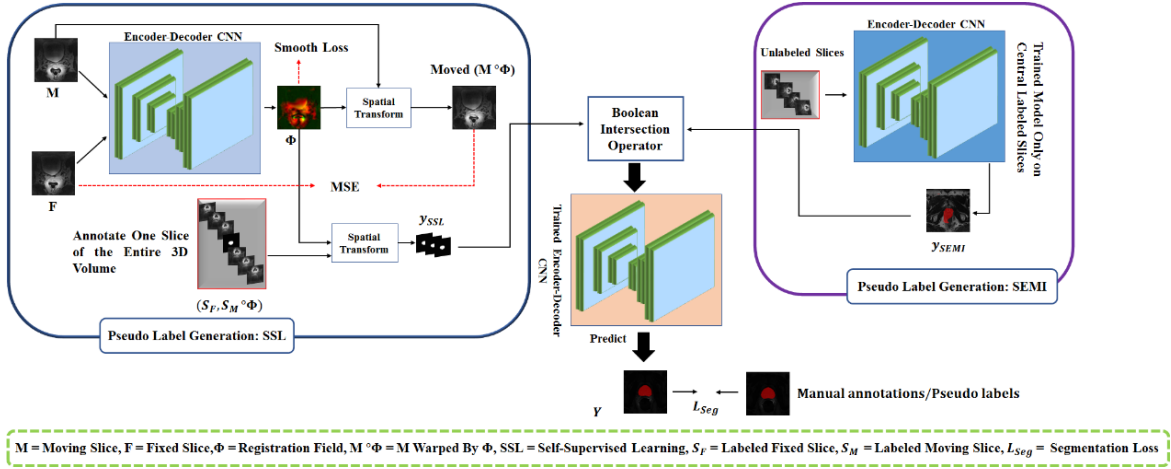

Figure 1. Overview of the proposed framework

## 2.1. Generating Pseudo Labels for Unannotated Slices in 3D

### 2.1.1. Pseudo label generation by semi-supervised learning

Let $D = \{D_1, D_2, D_3, ..., D_M\}$ represent the training 3D images, each of which contains N slices (i.e., 2D), $D_m = \{D_m^1, D_m^2, D_m^3, ..., D_m^N\}$ where $D_m^n \in \mathbb{R}^{H \times W}$. We suppose that experts need only to annotate the central slice of each volume. Accordingly, the set of ground truth masks that are available for model training set $Y = \{Y_1^{c1}, Y_2^{c2}, Y_3^{c3}, ..., Y_M^{cM}\}$ where $Y_m^{cn} \in \mathbb{R}^{H \times W}$ has the same size as its corresponding central slice, $D_m^{cn}$.

A segmentation model is firstly trained utilizing the labeled slices $D_L = \{(D_1^c, Y_1^c), (D_2^c, Y_2^c), ..., (D_M^c, Y_M^c)\}$ for certain epochs (we empirically set it to 50 epochs). Then, the unlabeled slices are introduced by using pseudo labels generated by the trained model (Figure 1). Target classes for unlabeled data that act as if they are real labels are known as pseudo-labels. Inspired by [6], for each unlabeled voxel, we simply select the class with the highest predicted probability as its pseudo label:

$$\bar{Y}_{m,k}^i = \begin{cases} 1 & if \ k = argmax_K f_{\bar{m}}(D_m^i) \\ 0 & otherwise \end{cases} \quad (1)$$

Here, $k$ refers to one of the segmentation classes and $K$ represents the different classes. A loss function combined of cross-entropy loss and dice loss. For both labeled and unlabeled slices, the same loss function is calculated. It needs to be mentioned that for unlabeled data, pseudo labels are regenerated after every network weight update training step.

### 2.2.2. Pseudo label generation by self-supervised learning

Transferring the labels from the labeled central slices to the corresponding unlabeled slices is another method of producing pseudo labels. We develop a self-supervised image registration technique to achieve this. Specifically, following our previous work [11], a deformable transformation network is optimized to perform slice-based image registration (Figure 1).

Suppose a pair of adjacent slices of 2D images in a 3D MR image are is represented as $(F, M)$. Here, $F$ represents the fixed slice and $M$ is the moving slice. The convolutional registration network takes the moving slices and the fixed two slices as input and computes a registration deformation field, $\Phi = Conv_\theta(F, M)$ from the moving slice to the fixed slice. Then, $\Phi$ is utilized to warp the moving slices and to get the moved slice, which is also the registered slices in 2D. It is worth noting that we only used image slices without their annotations are utilized during training the whole model training process. Therefore, the method is self-supervised.

Pseudo labels for unlabeled slices are generated by propagating the labels from the central slices to the boundary slices. We begin by treating the central slice of each image as the moving slice and their adjacent slices as fixed slices. After obtaining the deformation fields between the moving slice and the fixed slices using the deformable transformation network, pseudo labels of the adjacent slices can be obtained by warping the label of the central slice with the calculated deformation fields. This step is repeated until all the unlabeled slices are automatically labeled.

Unsupervised losses are utilized for training the registration model on all pairs of adjacent slices from 3D volumes. These losses comprise a smooth loss for local spatial variation in the deformation field ($\Phi$) and a similarity loss for variations in appearance between the warped image (M°$\Phi$) and the fixed image (F)

## 2.2. Collaborative Learning to Enhance the Accuracy of Generated Pseudo Labels

The two sets of pseudo labels generated by semi-supervised learning and self-supervised learning are fused in order to get more confident pseudo labels for the final target segmentation network optimization. Methods that use Boolean operations are straightforward and simple. We realize that intersection is a good solution in this case and that better pseudo labels could be generated with very little additional work:

$$y_{final} = y_{SEMI} \cap y_{SSL} \quad (2)$$

Where $y_{SEMI}$ and $y_{SSL}$ are the corresponding pseudo labels obtained by semi-supervised learning and self-supervised learning, respectively, and $y_{final}$ denotes the fused pseudo label. Target segmentation networks are trained using $y_{final}$.

## 2.3. Optimizing the Target Network for Volumetric MR Image Segmentation

Lastly, an encoder-decoder segmentation network is trained using both the manually labeled and pseudo labeled data to segment target objects. To train the network, a segmentation loss function combining dice loss and cross-entropy loss is calculated:

$$L_{Seg} = L_{Dice}(Y,F) + \gamma \cdot L_{CE}(Y,F) \quad (3)$$
$$Y = \begin{cases} y, & X \in \{D^c\} \\ y_{final}, & X \in \{D \setminus D^c\} \end{cases}$$

Where $F$ represents the prediction of the model, and $\gamma$ is a constant to balance the contributions of dice loss and cross-entropy loss. $y$ refers to the labels provided manually, and $y_{final}$ refers to the generated pseudo labels. $X$ represents the input image, $\{D^c\} = \{D_1^c, D_2^c, D_3^c, \ldots, D_M^c\}$ represents the central slice set, and $D \setminus D^c$ represents the slice set without the central slices. Only the final segmentation network is needed for inference.

## 3. EXPERIMENTS

To validate the effectiveness of the proposed framework, extensive experiments were conducted utilizing the MICCAI Prostate MR Image Segmentation (PROMISE12) challenge dataset [1]. In total, there are 50 3D MR images collected using a variety of equipment and protocols from multiple hospitals. Since the dataset is relatively small, 5-fold cross-validation experiments were conducted to avoid biased results caused by data split. The final step target object segmentation is carried out by a general and straightforward U-shaped network that has a contracting path encoder and an expansive path decoder [12]. The ADAM optimizer was employed to update the network parameters. The network was trained for 100 training epochs with a batch size of 4. The learning rate was initially set to 0.0001 and decayed by half every 30 epochs. The constant ($\gamma$) in Equation (3) was set to 1. Data augmenting strategies, including rotation and flipping were used to make better use of the limited training data.

Results of the proposed framework were compared to four semi-supervised learning approaches (mean teacher (MT) [13], interpolation consistency training (ICT) [14], deep co-training (DCT-Seg) [15], and uncertainty-aware mean teacher (UA-MT) [16]) to further assess the effectiveness of the proposed approach. Moreover, results from supervised counterpart, trained with only the labeled central slices (FS-LCS), was also reported.

Intersection-over-union (IoU), Dice score (Dice similarity coefficient, DSC), average symmetric surface distance (ASSD), and relative area/volume difference (RAVD) are calculated to quantitatively evaluate the segmentation performance. Better segmentation results are indicated by higher DSC and IoU values, as well as lower RAVD and ASSD values.

## 4. RESULTS

Table 1 lists the quantitative results of different methods. Unsurprisingly, FS-LCS generates the worst segmentation results among the six comparison methods. For the four semi-supervised learning methods (MT, ICT, DCT-Seg, and UA-MT), ICT behaves the best, followed by UA-MT and MT. DCT-Seg gives the lowest segmentation accuracy. Nevertheless, the Dice and IoU scores of DCT-Seg are still much higher (>7%) than those of FS-LCS, indicating the importance of information exploitation from unlabeled data.

Our proposed framework performs better on all the evaluation metrics than the four semi-supervised methods and FC-LCS by a large margin. Both the Dice score and IoU are increased by over 6% when compared to the best semi-supervised learning method (ICT). RAVD and ASSD are also largely reduced. These results confirm that our proposed collaborative framework is able to generate more confident pseudo labels and exploit the unlabeled data more effectively than the existing methods.

Figure 2 plots the segmentation maps of different methods for visual comparisons. Here, it is obvious that the segmentation maps generated by FS-LCS show large discrepancies from the reference (GT). The four semi-supervised methods and our method give more appealing prostate delineations that FS-LCS. Still, the results of our proposed method have higher overlap ratios with the reference (GT) than those of the four semi-supervised learning methods, demonstrating the effectiveness and robustness of our proposed framework.

Table 1. Quantitative results (Mean ±S.D.) on PROMISE12 validation set by 5-fold cross-validation experiments

| Methods | Dice [%] | IoU [%] | ASSD | RAVD |
|---|---|---|---|---|
| FS-LCS [12] | 63.4 ±18.6 | 48.6 ±16.9 | 5.08 ±3.24 | 0.3 ±0.66 |
| MT [13] | 73.1 ±11.6 | 58.9 ±13.4 | 5.76 ±5.40 | 0.041 ±0.38 |
| ICT [14] | 74.6 ±9.2 | 60.3 ±11.2 | 4.92 ±5.38 | 0.026 ±0.32 |
| DCT-Seg [15] | 70.8 ±12.1 | 56.0 ±13.5 | 8.06 ±7.95 | 0.357 ±0.51 |
| UA-MT [16] | 74.0 ±9.5 | 59.5 ±11.3 | 5.36 ±5.52 | 0.089 ±0.29 |
| **Ours** | **81.0 ±6.8** | **68.6 ±8.9** | **2.35 ±1** | **0.08 ±0.22** |

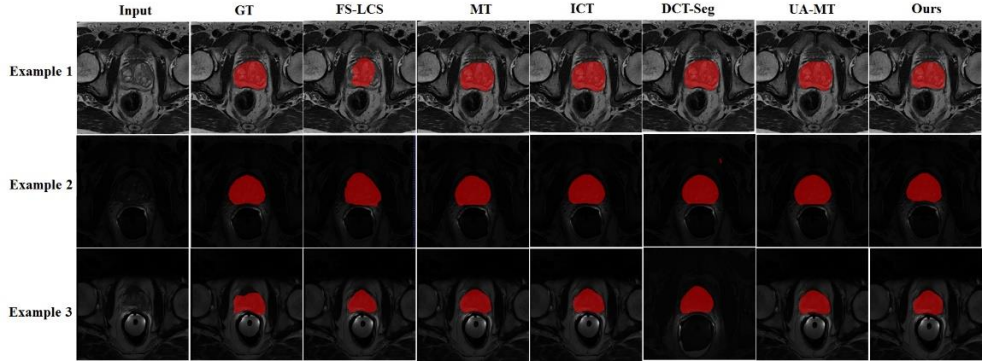

Figure 2. Examples prostate segmentation maps of different methods. Red color regions indicate the prostate regions segmented manually (GT, stands for ground truth) and automatically by different methods (FS-LCS to Ours).

## 5. CONCLUSIONS

In this study, we developed a semi-supervised and self-supervised collaborative learning framework for 3D prostate segmentation in MR images. In our framework, only the central slice of each 3D training image needed to be annotated manually, which can largely decrease the human effort spent on constructing the training dataset. To fully exploit the unlabeled image slices, pseudo labels were generated by two different methods, a semi-supervised learning method and a self-supervised learning method. Then, more confident pseudo labels were obtained by fusing the outputs of the two methods. Comprehensive comparisons between our proposed framework and the existing approaches were made utilizing an open-source prostate MR image dataset, and the results validated the effectiveness of our proposed framework. By largely decreasing the manpower on data annotating, our proposed framework can serve as a better baseline for clinical applications.

## 6. ACKNOWLEDGMENTS

This research was partly supported by Scientific and Technical Innovation 2030-"New Generation Artificial Intelligence" Project (2020AAA0104100, 2020AAA0104105), the National Natural Science Foundation of China (61871371), Guangdong Provincial Key Laboratory of Artificial Intelligence in Medical Image Analysis and Application (2022B1212010011), the Basic Research Program of Shenzhen (JCYJ20180507182400762), Shenzhen Science and Technology Program (RCYX20210706092104034), and Youth Innovation Promotion Association Program of Chinese Academy of Sciences (2019351).